\documentclass[graybox]{svmult}

\usepackage{type1cm}        
\usepackage{makeidx}         
\usepackage{graphicx}        
\usepackage{multicol}        
\usepackage[bottom]{footmisc}

\usepackage{newtxtext}       %
\usepackage{newtxmath}       

\usepackage{array}
\usepackage{multirow}
\usepackage{acronym}

\newcommand{\etal}{\textit{et al.}}

\makeindex             

\begin{document}

\title*{Model-Based Reinforcement Learning for Type 1 Diabetes Blood Glucose Control}
\author{Taku Yamagata, Aisling O'Kane, Amid Ayobi, Dmitri Katz, Katarzyna Stawarz, Paul Marshall, Peter Flach and Ra{\'u}l Santos-Rodr{\'i}guez}
\authorrunning{T.Yamagata \etal}
\institute{Taku Yamagata \at University of Bristol, Bristol BS8 1UB UK \email{taku.yamagata@bristol.ac.uk}
\and Amid Ayobi \at University of Bristol, Bristol BS1 5DD UK \email{amid.ayobi@bristol.ac.uk}
\and Aisling O’Kane \at University of Bristol, Bristol BS8 1UB UK \email{a.okane@bristol.ac.uk}
\and Dmitri Katz \at The Open University, Milton Keynes MK7 6AA UK \email{dmitri.katz@open.ac.uk}
\and Katarzyna Stawarz \at Cardiff University, Cardiff CF24 3AA UK \email{StawarzK@cardiff.ac.uk}
\and Paul Marshall \at University of Bristol, Bristol BS8 1UB UK \email{p.marshall@bristol.ac.uk}
\and Peter Flach \at University of Bristol, Bristol BS8 1UB UK \email{Peter.Flach@bristol.ac.uk}
\and Ra{\'u}l Santos-Rodr{\'i}guez \at University of Bristol, Bristol BS8 1UB UK \email{enrsr@bristol.ac.uk}
}
%
%
\maketitle

\abstract{In this paper we investigate the use of model-based reinforcement learning to assist people with Type 1 Diabetes with insulin dose decisions. The proposed architecture consists of multiple Echo State Networks to predict blood glucose levels combined with Model Predictive Controller for planning. Echo State Network is a version of recurrent neural networks which allows us to learn long term dependencies in the input of time series data in an online manner. Additionally, we address the quantification of uncertainty for a more robust control. Here, we used ensembles of Echo State Networks to capture model (epistemic) uncertainty.
We evaluated the approach with the FDA-approved UVa/Padova Type 1 Diabetes simulator and compared the results against baseline algorithms such as Basal-Bolus controller and Deep Q-learning. The results suggest that the model-based reinforcement learning algorithm can perform equally or better than the baseline algorithms for the majority of virtual Type 1 Diabetes person profiles tested.}

\section{Introduction}
\label{sec:1}
Type 1 Diabetes is a chronic condition that is characterized by the lack of insulin secretion and resulting in uncontrolled blood glucose level increase \cite{Alberti1998,Davis2015}.  High blood glucose levels for extended periods of time can result in permanent damage to the  eyes, nerves, kidneys and blood vessels, while low blood glucose levels can lead to death \cite{Mol2004,Mynatt2010,NHS}. To manage blood glucose level, people on multi-dose injection (MDI) therapy usually take two types of insulin injections: basal and bolus. The basal is long-acting insulin, which provides a constant supply of insulin over 24-48 hours, helping maintain resting blood glucose levels. The bolus is fast-acting insulin which helps to suppress the peak of the blood glucose levels caused by meals or to counteract hyperglycemia \cite{NHS}. People with diabetes must make constant decisions of the timing and amount of these insulin injections, which is often challenging as insulin requirements for meals can change depending upon many factors such as exercise, sleep, or stress. The idiosyncratic nature of the condition means that triggers, symptoms and even treatments are often quite individual \cite{Mianowska2011,Okane2016a,Okane2016b,Pesl2017,Storni2011}, which creates challenges to developing diabetes self-management technologies. 

\begin{figure}[h]
\sidecaption
\includegraphics[scale=.35]{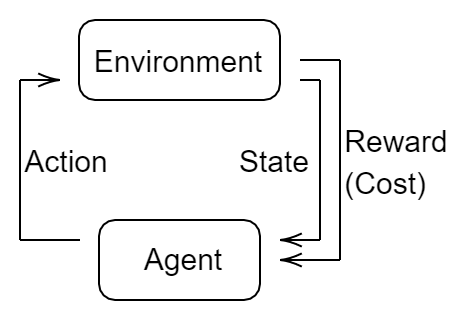}
\caption{Reinforcement learning framework overview.}
\label{fig:RL}    
\end{figure}

In this paper we consider the benefits of using \ac{MBRL} to assist decisions about bolus insulin injections. The goal of \ac{RL} is to learn sequences of actions in an unknown environment~\cite{Sutton1998}. The learner (Agent) interacts with the environment, observes its consequences, and  receives a reward (or a cost) signal, which is a numerical number assessing current the situation. The agent decides a sequence of actions to maximize the reward (or minimize the cost) as shown in Fig.\ref{fig:RL}. \ac{RL} is well-suited to this task because it can learn the model in an online manner with minimal assumptions about the underlying process of the blood glucose behaviour and hence can adapt to different individuals or changes over time. \ac{MBRL} is particularly well suited to this objective because it is more sample-efficient than alternative \ac{RL} approaches (\ac{MFRL}) and also allows us to generate predictions for consequences of counterfactual actions that can be used as \emph{explanations} of the suggestion. In our \ac{MBRL} setting, we also can estimate the confidence level of the predictions by using the prediction uncertainty. It is very important to show the explanation for the suggestion together with its confidence level so that the person that receives the suggestion can make a decision whether they would follow the recommended course of action.

As a first step towards realising such a recommender system, we investigated how well \ac{MBRL} can learn the insulin injection decision and compared it with both a typical \ac{MFRL} algorithm (\ac{DQN}) and an algorithm that mimics human decision-making (\ac{BBController}). We used an FDA-approved Type 1 Diabetes computer simulator and let the algorithms decide the insulin injections and evaluated its blood glucose level behaviours.

 Our \ac{MBRL} approach builds upon previous work on \acp{ESN} \cite{Jaeger2010,Gurbilek2013}, the ensembles of models for \ac{MBRL} \cite{Chua2018} and \ac{MPC} for artificial pancreas \cite{Cameron2011, Bequette2013}. However we believe this is the first attempt to combine these algorithms for the Type 1 Diabetes blood glucose level control task, and evaluate its performance against non-\ac{MBRL} algorithms.

This paper is organized as follows. Section \ref{sec:related_work} introduces related work regarding the blood glucose control task. Section \ref{sec:methods} describes our \ac{MBRL} method. Section \ref{sec:evaluation} presents our evaluation method, benchmark algorithms and the evaluation results. Finally, Section \ref{sec:further_work} concludes with a summary and possible future work.
\section{Related Work}
\label{sec:related_work}
Several attempts have been made for a closed-loop artificial pancreas, especially in the control system society using \ac{MPC}~\cite{Bequette2013}, proportional-integral-derivative control~\cite{Steil2013} and fuzzy logic~\cite{Atlas2010}.

However, there are relatively few studies on the blood glucose levels control task using \ac{RL} approaches. Most of the early works employ compartmental blood glucose and insulin models to infer some of insulin/glucose related internal states of human body, and then learn its insulin injection policy with relatively simple \ac{MFRL} algorithms such as Q-Learning~\cite{ngo2018ff,ngo2018} or Actor-Critic~\cite{daskalaki2013,Daskalaki2013a}.
Fox \etal\ employed more recent \ac{RL} techniques~\cite{Fox2019}, such as deep neural networks for the Q-Learning algorithm -- arguably the most common \ac{MFRL} algorithm. They showed that although the agent was not given any prior knowledge of the blood glucose/insulin relations, it learns its insulin injection policy and achieves performance comparable with existing algorithms.

In the field of model-based system control several approaches exist -- we refer the reader to~\cite{Bequette2013} and the references therein. The closest to our work is~\cite{Cameron2011}, where the authors use a linear compartmental model for predicting the mean and variance of the future blood glucose levels. It exploits \ac{MPC} for planning by taking into account the variance of the blood glucose level prediction. The main differences from our work are: (1) they employ a linear compartmental model which has a small number of parameters and hence easier to learn, whereas we use more generic recurrent neural networks, which have greater flexibility to adapt to any personal blood glucose level behaviour; (2) their model parameters are learnt off-line, whereas ours are adjusted  online; and (3) the handling of uncertainty -- we measure the model's uncertainty while they measure the uncertainty involved in meal events.

\section{Methods}
\label{sec:methods}
In order to apply \ac{RL} algorithms to this problem, we formulate the task as Markov Decision Process (MDP), which has four tuples $(S,A,p,c)$ where $S$ is a set of states, $A$ is a set of actions, $p$ is the state transition probabilities and $c$ is a cost function. Essentially the blood glucose control task is a Partially Observable MDP, however we see it as an MDP by defining state $S$ as all history of insulin doses and carbohydrate intakes. 

More precisely, the overall pipeline makes use of \acp{ESN}  to store the history in its hidden states, shown in Section\ref{subsec:ESNs}. The corresponding actions $A$ are the dosages of bolus insulin. We exploit the risk function introduced in \cite{Kovatchev1997} as  our cost function $c$, described in Section~\ref{subsec:cost}. While we use the \acf{MBRL} algorithm with \acp{ESN} for the prediction of blood glucose levels, \ac{MPC} generates the insulin dose suggestions from the blood glucose level predictions (Section~\ref{subsec:MPC}) and their uncertainty estimations (Section~\ref{subsec:uncertainty}).
.

\subsection{Cost function}
\label{subsec:cost}
\begin{figure}[b]
\sidecaption
\includegraphics[scale=.53]{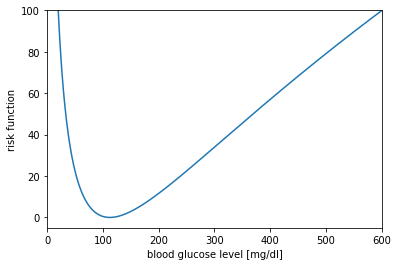}
\caption{Risk function proposed by Kovatchev \etal\ \cite{Kovatchev1997} The figure shows the relationship between blood glucose level [mg/dl] and its risk function value. The risk function value shows asymmetric shape -- increase rapidly for low blood glucose levels compare to the high blood glucose levels, which is aligned with the clinical risk of low and high blood glucose levels.}
\label{fig:risk}       
\end{figure}

For our task, it is natural to use as cost function a measure of risk associated with the given blood glucose level. However it is not straightforward to define such a measure, as it presents different scales of risks between higher than normal blood glucose levels (hyperglycemia) and lower than normal blood glucose levels (hypoglycemia). Kovatchev \etal\ proposed the following expression to symmetrize the risks of hyper and hypoglycemia~\cite{Kovatchev1997}. This blood glucose risk function $f_r$ is defined as in Eq.~\ref{eq:risk}. The blood glucose level transition from 180 to 250mg/dl would appear threefold larger than a transition from 70 to 50mg/dl, whereas these are similar in terms of the risk function variations.
\begin{equation}
f_r(BGL) =  15.09 \cdot \left(log(BGL)^{1.084} - 5.381\right)^2
\label{eq:risk}
\end{equation}
where BGL is the blood glucose level in mg/dl. Fig.~\ref{fig:risk} shows the mapping between blood glucose level (x-axis) to the risk function (y-axis).
We used the risk function value as the cost function, hence our \ac{RL} agent searches a policy minimising the total risk values over an episode. 


\subsection{Echo State Networks}
\label{subsec:ESNs}
\acp{ESN} were proposed as an alternative structure of standard recurrent neural networks in machine learning~\cite{Jaeger2010}. They are also called liquid state machine in computational neuroscience~\cite{Gurbilek2013}.  \acp{ESN} take an input sequence $\mathbf{u}=(\mathbf{u}(1), \mathbf{u}(2), ..., \mathbf{u}(T))$ by recursively processing each symbol while maintaining its internal hidden state $\mathbf{x}$. At each time step $t$, the \ac{ESN} takes input $\mathbf{u(t)} \in \mathbb{R}^K$ and updates its hidden state $\mathbf{x}(t)\in\mathbb{R}^N$ by:
\begin{eqnarray}
    \tilde{\mathbf{x}}(t) &= f(\mathbf{W}^{in} \cdot \mathbf{u}(t) 
    + \mathbf{W} \cdot \mathbf{x}(t-1)) \\
    \mathbf{x}(t) &= (1-\alpha) \cdot \mathbf{x}(t-1) + \alpha \cdot \tilde{\mathbf{x}}(t),
\end{eqnarray}
 where $f$ is the internal unit activation function, which is $tanh$ in our model, $\mathbf{W}^{in}\in\mathbb{R}^{N\times K}$ is the input weight matrix, $\mathbf{W}\in\mathbb{R}^{N\times N}$ is the internal connections weight matrix and $\alpha \in (0,1]$ is the leakage rate, which controls the speed of the hidden states change hence controls the output smoothness.
 
 The output at time step $t$, $\mathbf{y} (t)\in\mathbb{R}^L$ is obtained from the hidden states and the inputs by:
 \begin{equation}
    \mathbf{y}(t) = f^{out}\left(\mathbf{W}^{out} \cdot \left[ \mathbf{x}(t)^T, \mathbf{u}(t)^T \right]^T\right),
    \label{eq:ESN_y}
\end{equation}
 where $f^{out}$ is the output unit activation function (which is the identity function in our model as we are dealing with a regression task) and  $\mathbf{W}^{out}\in\mathbb{R}^{L\times (N+K)}$ is the output weights matrix.
 
 The matrices for updating the hidden states, $\mathbf{W}^{in}$ and $\mathbf{W}$, are randomly initialized and fixed (not updated during learning process), only the output weights matrix $\mathbf{W}^{out}$ is leaned to obtain the target output sequences. As it only learns the output weights, it doesn't require back propagation through the network nor time, hence it learns much faster than the normal recurrent neural networks. The downside of using \ac{ESN} is that it requires much higher number of hidden states to achieve good performance, hence it required more computational power for inference. 
 
 To make \acp{ESN} work properly, the fixed weights must satisfy the so-called \emph{echo state property}: the internal states $\mathbf{x}(t)$ should be uniquely defined only by the past inputs $\mathbf{u}(k)|_{k=...t}$~\cite{Jaeger2010}. The actual method to initialise the weights can be found in~\cite{Luko2012}, which also gives useful guidance for using \acp{ESN}.

\subparagraph{\acp{ESN} for the blood glucose level prediction task}
In our work, the \ac{ESN} takes a sequence of bolus insulin injection and carbohydrate intakes as inputs, and predicts the blood glucose level. 

To learn the \ac{ESN} output weights we use the Mean Squared Error between predicted and observed blood glucose levels as loss function.
\begin{equation}
   \mathcal{L}_{d}(\theta)=\frac{1}{T}\sum_{t=1}^{T}(\mu_\theta(t) - BGL(t))^2
    \label{eq:LOSS_D}
\end{equation}
Here, $\mu _\theta(t)$ is the predicted blood glucose level by \ac{ESN} at time step $t$, where $\theta$ is the optimization parameter (here it is $\mathbf{W}^{out}$) and  $BGL(t)$ is observed blood glucose level. As it can be seen as a linear regression problem, the output weights are derived by solving the Normal equation~\cite{Luko2012}. 

To capture model (epistemic) uncertainty, it applies multiple instances of \acp{ESN}, and each of them has different input and internal connection weights. \acp{ESN} are well suited for the ensemble approach as it has fixed random internal weights which project the inputs sequence into different hidden states. So naturally they output different values where there is no training data, capturing higher epistemic uncertainty. In our evaluation, we employ five instances of ESNs, which is suggested by \cite{Chua2018}.


\subsection{Uncertainty quantification}
\label{subsec:uncertainty}
We employ multiple \acp{ESN} to capture the uncertainty in predicted blood glucose level. 
They produce multiple predictions of the blood glucose levels from the \ac{ESN} models for each action sequence. To quantify the cost (risk) of uncertainty, we take the mean of the cost of the predicted blood glucose levels for each of action sequence $\frac{1}{MT}\sum_{t=n}^{n+T-1} \sum_{m=1}^{M} c(BGL_t^m)$, where $c(.)$ is a cost function, $BGL_t^m$ is blood glucose levels prediction from \ac{ESN} model $m$ at time step t, and $M$ and $T$ are number of \ac{ESN} models and number of time steps in the action sequence. We then select the action sequence which minimises this mean cost.

We encourage (optimistic or exploratory approach) or discourage (pessimistic or safe approach) taking risks by designing the cost function accordingly. Here we define a risk margin $RM$ as the difference between the averaged cost function and cost of the averaged blood glucose level predictions.
\begin{equation}
    RM = E[c(BGL)] - c(E[BGL]).
    \label{eq:RiskMargin}
\end{equation}
A positive (negative) risk margin means our metric $E[c(BGL)]$ discourages (encourages) taking risks. If we use a convex cost function as described in Section~\ref{subsec:cost}, $RM$ is positive according to Jensen's inequality, hence it discourages risks.


\subsection{Model Predictive Controller}
\label{subsec:MPC}
Model predictive controller (MPC) is a planning method to facilitate control of systems with a long time delay and non-linear characteristics. The \ac{MPC} uses a prediction model to estimate the consequences of a sequence of actions and repeats the process for many action sequences. Then it picks the sequence of actions that gives the best consequence and applies the first action of the sequence. In the next time step this process is repeated. This effectively means it re-plans the sequence of actions based on the latest state information from the environment, which makes the algorithm robust against any noise or prediction errors.

There are several algorithms to generate the sequence of actions to test -- such as random shooting~\cite{Rao2010} and cross entropy method~\cite{Boer2005}. In our work, we use a fixed table for the sequence of actions to test. The table has six action sequences, each of which takes a different amount of bolus injection as its first action. The amount of bolus injection at the first action is \{0, 5, 10, 20, 40, 80\} times of the person's basal infusion rate. Following the approach of~\cite{Fox2019}, the basal infusion rate is given for each virtual person's model, and we use it to scale the bolus injection. While our model generates suggestions for bolus injections, for the basal injections, it assumes the person is taking the given basal infusion rate.
The action sequence length (time horizon) is set to 48 time steps, which is 4 hours long as each time step represents a five-minute period. Each action sequence has a bolus injection as the first action of the sequence. We believe this is sensible because the bolus injections is normally taken just after or before a meal and there is no meal announcement in our system at moment (the algorithm does not know the meal event until it happens). Therefore, the best time to take bolus injection would be immediately after detecting the meal event, which is the first action in the sequence.
A proper meal announcement mechanism is left for future work.

\section{Evaluation}
\label{sec:evaluation}
\begin{figure}[b]
\sidecaption
\includegraphics[scale=.25]{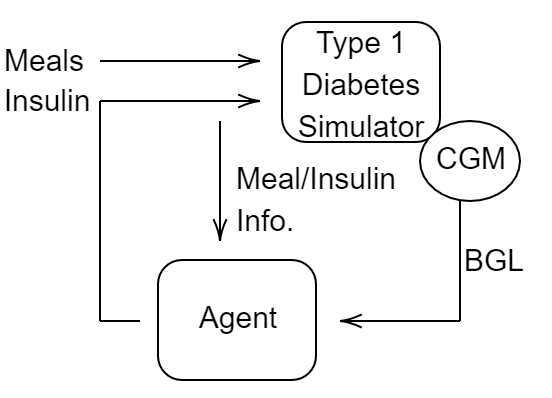}
\caption{Evaluation system top level diagram.}
\label{fig:eval}       
\end{figure}

We empirically evaluated how well the \acf{MBRL} can learn insulin injection decisions and compared it with a typical \acf{MFRL} algorithm and also with a non-\ac{RL} algorithm designed to mimic human decision-making. In this paper, we did not compare the blood glucose level prediction accuracy with other prediction models. Instead, we focused on evaluating the performance of the agents. The overview of the evaluation system is shown in Fig.~\ref{fig:eval}. We used an FDA-approved Type 1 Diabetes simulator, which takes meal and insulin injection information, then outputs a blood glucose level (BGL) as a \ac{CGM} reading at each time step. The algorithms (agents) receive the meal, insulin and blood glucose level information and decides the amount of insulin taking in the next time step. We simulated the algorithms together with the Type 1 Diabetes simulator, and evaluated how well the blood glucose levels are managed.

\subsection{UVa/Padova Type 1 Diabetes simulator}
The UVa/Padova Type 1 Diabetes Simulator~\cite{DallaMan2014} was the first computer model accepted by the FDA as a substitute for preclinical trials of certain insulin treatments, including closed-loop algorithms. The model takes carbohydrate intakes and insulin injection as inputs, simulates human body insulin/blood glucose behaviours and outputs the blood glucose level measurements. It has gastro-interstinal tract, glucose kinetics and insulin kinetics sub models. Each of these sub models is defined with differential equations with parameters to simulate different individuals. 
Our simulator is based on an open source implementation of the UVa/Padova Type 1 Diabetes simulator~\cite{Xie2018}, which comes with different profiles for 30 virtual people with type 1 diabetes -- ten each for children, adolescents and adults. Our experiments use nine virtual people, three of each age group.

\subsection{Benchmark algorithms}
We used two benchmark algorithms to compare the proposed approach against, one from \ac{RL} algorithms (GRU-DQN) and the other one from non-\ac{RL} approaches (\ac{BBController}). These are described below. 

\subparagraph{GRU-DQN}
Deep Q-Learning (DQN) is a common \ac{MFRL} algorithm, which learns the action-value function $Q(s,a)$ -- expected cumulative future rewards starting with state $s$ and action $a$. It then uses the learned action value function to decide which action to take at time step $t$ by $a_t=argmax_{a\in\mathcal{A}}Q(s_t,a)$. In our work, the agent observes the blood glucose levels from a \ac{CGM}, carbohydrate intakes and insulin injections, and infers the action value function. It is a partially observable model so we used \ac{GRU} to infer the hidden states and approximate the action value function. GRU-DQN was successfully applied to this problem before~\cite{Fox2019} so we followed their same set up which involves two \ac{GRU} recurrent layers of 128 hidden states and followed by a fully connected output layer size of 128. However, our our states (the input of GRU-DQN) include carbohydrate information, whereas \cite{Fox2019} does not. We include it here to make our comparison fair against the \ac{MBRL} algorithm, which has acess to the carbohydrate information.  

\subparagraph{BBController}
Basal-Bolus Controller mimics how an individual with Type 1 Diabetes controls their blood glucose levels. The UVa/Padova simulator comes with the necessary parameters for this algorithm for each of the virtual people with Type 1 Diabetes models, such as basal insulin rates $\mathnormal{bas}$, a correction factor $\mathnormal{CF}$ and a carbohydrate ratio $\mathnormal{CR}$. The simulator decides the amount of insulin injection by $\mathnormal{bas}+(c_t>0)\cdot(c_t/\mathnormal{CR} + (b_t > 150) \cdot (b_t - b_{tgt})/\mathnormal{CF})$, where $c_t$ is carbohydrate intake at time step $t$, $b_t$ is the blood glucose measurements, $b_{tgt}$ is a target blood glucose level. The last term is only applied when the blood glucose measurement exceeds 150 mg/dl. We use the implemented model that comes with the Type 1 Diabetes simulator~\cite{Xie2018}.

\subsection{Simulation Conditions}
Each episode lasts 24 hours, starting at 6am and finishing at 6am the next day. Three meals and three snack events are simulated with some randomness in terms of amount, timing and also whether they take the meal/snack. The timing follows a truncated normal distribution and the amount is normally distributed. The meal parameters are shown in Table~\ref{tab:meal}.
\begin{table}[!t]
\caption{Parameters for meal event generator.}
\label{tab:meal}       
%
%
\newcolumntype{C}[1]{>{\hfil}m{#1}<{\hfil}}
\begin{tabular}{p{1.5cm}  C{1.2cm} C{1.2cm} C{1.2cm}  C{1.2cm}  C{1.2cm} C{1.2cm}  C{1.2cm} }
\hline\noalign{\smallskip}
\multicolumn{1}{c|}{} & \multicolumn{1}{c|}{} & \multicolumn{4}{c|}{Time [hours]} & \multicolumn{2}{c}{Carbs. [g]}  \\
\multicolumn{1}{l|}{Meal type} &  \multicolumn{1}{c|}{Prob.} & \multicolumn{1}{c}{lower} & \multicolumn{1}{c}{upper} & \multicolumn{1}{c}{mean} & \multicolumn{1}{c|}{std.} & \multicolumn{1}{c}{mean} & \multicolumn{1}{c}{std.}              \\
\multicolumn{1}{c|}{} &   \multicolumn{1}{c|}{} & \multicolumn{1}{c}{bound} & \multicolumn{1}{c}{bound} &  & \multicolumn{1}{c|}{} &  &               \\
 
\noalign{\smallskip}\svhline\noalign{\smallskip}
Breakfast & 0.95 & 5 & 9    & 7   & 1   & 45 & 10\\
Snack\#1  & 0.3  & 9 &  10 & 9.5 &  0.5 &  10 &  5\\ 
Lunch       & 0.95 & 10&  14 & 12  &  1   &  70 &  10\\
Snack\#2  &  0.3 & 14&  16 & 15  &  0.5 &  10 &  5\\
Dinner      & 0.95 & 16&  20 & 18  &  1   &  80 &  10\\
Snack\#3  &  0.3 & 20&  23 & 21.5&  0.5 &  10 &  5\\
\noalign{\smallskip}\hline\noalign{\smallskip}
\end{tabular}
\end{table}
The agent receives information from the environment such as the meal (carbohydrate), insulin and blood glucose levels, and decides the insulin dose for the next time step. Each time step is set to five minutes in length. In this evaluation, the person does not take food to compensate for low blood glucose levels (the meal event always follows a pre-defined order as described above). While this is not realistic, it is a good way to measure how well the algorithm works because ultimately we would like to develop an algorithm that does not require any corrections from the user. The episode is terminated if the blood glucose level goes below 20 mg/dl or beyond 600 mg/dl, as these limit are extreme and they are outside of the possible blood glucose level range considered by~\cite{Kovatchev1997}.

\subsection{Results}
We train \ac{MBRL} for 200 episodes and GRU-DQN for 1000 episodes, then use the last 30 episodes to measure the percentage of episodes completed without termination due to extreme blood glucose levels. For \ac{BBController}, we just run 30 episodes to measure, as it has pre-optimized model parameters and no training is required.

The results are given in Table~\ref{tab:1}. \ac{MBRL} gives better results than GRU-DQN and comparable with \ac{BBController}. \ac{MBRL} struggles with child\#002, \#003 and adolescent\#002. By looking into these cases, we found that \ac{MBRL} fails due to the \ac{MPC} time horizon not being long enough. The \ac{MPC} time horizon is set to 4 hours, hence the agent could not foresee a possible hypoglycemia event in the early morning after the person takes an evening meal. The agent suggests too much insulin, and it causes hypoglycemia in the early morning. This can be fixed by increasing the \ac{MPC} time horizon, but requires some additional consideration as it might lead to inappropriate suggestions during the day.
\begin{table}[!t]
\caption{\% of number of completed episodes without termination due to extreme blood glucose level value}
\label{tab:1}       
\newcolumntype{C}[1]{>{\hfil}m{#1}<{\hfil}}
\begin{tabular}{p{3cm} C{2.2cm} C{2.2cm} C{2.2cm}}
\hline\noalign{\smallskip}
Virtual Person Profile & BBContoller & GRU-DQN & MBRL  \\
\noalign{\smallskip}\svhline\noalign{\smallskip}
child\#001      &  30.0 &  3.3 & \textbf{100} \\
child\#002      & \textbf{90} & 23.3 & 53.3   \\
child\#003      & \textbf{66.7} & 43.3 & 30.0 \\
\noalign{\smallskip}
adolescent\#001 & \textbf{100} & \textbf{100} & \textbf{100} \\ 
adolescent\#002 & \textbf{66.7}  & 56.7 & 0.0      \\
adolescent\#003 & 90 & 20 & \textbf{100}  \\
\noalign{\smallskip}
adult\#001      &  \textbf{100} & 70.0 & 96.7 \\
adult\#002      &  \textbf{100} & \textbf{100} & \textbf{100} \\
adult\#003      &  96.7 & 16.7 & \textbf{100} \\
\noalign{\smallskip}\hline\noalign{\smallskip}
\end{tabular}
\end{table}

Table~\ref{tab:2} shows the percentage of time spent in a target blood glucose level range (70-180 mg/dl.) These are measured in the last 10 of the completed episodes(i.e., not terminated). Here \ac{MBRL} gives the best overall results compared to the other agents. Note that no data is available for adolescent\#002, as it fails to get any non-terminated episode (due to the reason described above).

\begin{table}[!t]
\caption{\% of time spent in the target blood glucose level range (70 - 180 mg/dl)}
\label{tab:2}       
\newcolumntype{C}[1]{>{\hfil}m{#1}<{\hfil}}
\begin{tabular}{p{3cm} C{2.2cm} C{2.2cm} C{2.2cm}}
\hline\noalign{\smallskip}
Virtual Person Profile & BBContoller & GRU-DQN & MBRL\\
\noalign{\smallskip}\svhline\noalign{\smallskip}
child\#001      &  44.0 & 28.3 & \textbf{59.6}\\
child\#002      & 42.6  & 38.2 & \textbf{55.3}\\
child\#003      & 40.7 & 36.0 & \textbf{45.1} \\
\noalign{\smallskip}
adolescent\#001 & 85.8 & 81.4 & \textbf{100.0}\\ 
adolescent\#002 & \textbf{49.0} & 39.8  & n/a \\
adolescent\#003 & 46.7 & 42.4 & \textbf{66.1} \\
\noalign{\smallskip}
adult\#001      &  \textbf{60.1} & 50.3 & 56.8 \\
adult\#002      &  \textbf{73.3} & 66.9 & \textbf{73.3} \\
adult\#003      &  58.7 & 46.9 & \textbf{68.8} \\
\noalign{\smallskip}\hline\noalign{\smallskip}
\end{tabular}
\end{table}

We also evaluated the effect of the uncertainty estimation by comparing the results from \ac{MBRL} with/without it. For \ac{MBRL} without uncertainty, we take an average over multiple \acp{ESN} predictions to come up with a single blood glucose prediction, and then we calculate its cost. Whereas \ac{MBRL} with uncertainty computes the cost of the all predictions, then takes average of the costs as described in Section~\ref{subsec:uncertainty}.

\begin{figure}[b]
\sidecaption
\includegraphics[scale=.55]{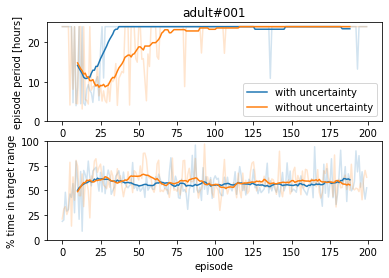}
\caption{Comparison between \ac{MBRL} with uncertainty and without uncertainty models. The upper plot shows the learning curve for simulated period for each episode, which goes up to 24 hours if the blood glucose level is controlled well. 
The lower plot shows \% of time spent in the target blood glucose range (70-180mg/dl)}
\label{fig:RiskNormalPlot}    
\end{figure}

Figure~\ref{fig:RiskNormalPlot} shows the learning curves for these two \ac{MBRL} algorithms with adult\#001. The upper plot shows the episode period, which goes up to 24 hours if there is no termination, and the bottom plot shows \% of time spent in the target blood glucose range. From the upper plot, the algorithm with uncertainty achieves ``no episode termination'' (24 hours episode) much earlier than the one without estimating uncertainty. At an early stage of the learning process, the prediction model is not very accurate, so it is much better by taking into account its uncertainty. For the later stages, the predictions become more accurate, hence it shows similar performance in both cases. Table~\ref{tab:3} shows asymptotic results of the percentage of time spent in the target blood glucose range, indicating that both have similar asymptotic performances.

\begin{table}[!t]
\caption{\% of time spent in the target blood glucose range (70 - 180 mg/dl)}
\label{tab:3}       
\newcolumntype{C}[1]{>{\hfil}m{#1}<{\hfil}}
\begin{tabular}{p{3cm} C{3cm} C{3cm}}
\hline\noalign{\smallskip}
Virtual Person Profile & MBRL & MBRL \\
                           & (with uncertainty) &  (without uncertainty) \\
\noalign{\smallskip}\svhline\noalign{\smallskip}
child\#001      &  59.6 &  57.5 \\
adolescent\#001 & 100.0 &  95.9 \\ 
adult\#001      &  56.8 &  56.7 \\
\noalign{\smallskip}\hline\noalign{\smallskip}
\end{tabular}
\end{table}


\section{Conclusions and Future Work}
\label{sec:further_work}

We investigated the use of \ac{MBRL} to assist Type 1 Diabetes decision-making by evaluating \ac{MBRL} with the FDA-approved UVa/Padova simulator. We compared the results with two baseline algorithms, GRU-DQN and \ac{BBController}. The results suggest that the \ac{MBRL} approach works better than the GRU-DQN algorithm and similar or slightly better than the \ac{BBController}. 
Also, our results show that taking into account the model uncertainty improves its performance in the early stages of learning.

There are several avenues for future work. At the present stage we only tested our algorithms with the UVa/Padova Type 1 Diabetes simulator, which is good for single meal scenarios but not for multiple meals~\cite{DallaMan2014}. This is primarily because the model has fixed parameters for each person and does not simulate meal-by-meal nor day-by-day parameter drifting. In addition, our current learning method must be extended to adapt to parameter drifts. A possible approach for such an extension would be to introduce meta-learning~\cite{Finn2017}.

Another area for further work relates to meal information. We assumed all meal events are correctly given by the person when the event is happening; however, this may not be very realistic as it is a considerable burden for a person to put every single meal event into the algorithm. It is also hard to know the exact carbohydrate count of each meal. Some researchers therefore structure the blood glucose predictor without having a meal input. Another alternative would be to have a model to back-predict a meal event from the observed blood glucose levels. We think it is possible to learn the meal event in conjunction with the blood glucose level prediction model with occasional human inputs.




\begin{acknowledgement}
This project is funded by the Innovate UK Digital Catalyst Award – Digital Health and is in partnership with Quin Technology.
\end{acknowledgement}


\bibliographystyle{spmpsci}
\bibliography{my_library}

\acrodef{RL}{reinforcement learning}
\acrodef{MBRL}{model-based reinforcement learning}
\acrodef{MFRL}{model-free reinforcement learning}
\acrodef{ESN}{Echo State Network}
\acrodef{MPC}{model predictive controller}
\acrodef{GRU}{gated recurrent units}
\acrodef{DQN}{deep Q-Learning}
\acrodef{CGM}{continuous glucose monitor}
\acrodef{BBController}{Basal-Bolus controller}

\end{document}